\documentclass[letterpaper]{article}
\usepackage{aaai2026} 
\usepackage{times} 
\usepackage{helvet} 
\usepackage{courier} 
\usepackage[hyphens]{url} 
\usepackage{graphicx} 
\usepackage{natbib}
\usepackage{caption}
\usepackage{algorithm}
\usepackage{algorithmic}
\usepackage{amsmath}
\usepackage{multirow}
\usepackage{enumitem} 
\usepackage{makecell}

\usepackage{subcaption} 
\usepackage{newfloat}
\usepackage{listings}
\DeclareCaptionStyle{ruled}{labelfont=normalfont,labelsep=colon,strut=off} 
\floatstyle{ruled}
\newfloat{listing}{tb}{lst}{}
\floatname{listing}{Listing}
\nocopyright
\usepackage{booktabs}
\title{MSNav: Zero-Shot Vision-and-Language Navigation with Dynamic Memory and LLM Spatial Reasoning}
\author{
  Chenghao Liu\textsuperscript{\rm 1}\thanks{These authors contributed equally to this work, ordered by randomly dice rolling.},
  Zhimu Zhou\textsuperscript{\rm 1}\footnotemark[1],
  Jiachen Zhang\textsuperscript{\rm 1},
  Minghao Zhang\textsuperscript{\rm 2},
  Songfang Huang\textsuperscript{\rm 1}\thanks{Corresponding author: hsf@pku.edu.cn},
  Huiling Duan\textsuperscript{\rm 1}
}
\affiliations{
    \textsuperscript{\rm 1}School of Advanced Manufacturing and Robotics, Peking University,
    \textsuperscript{\rm 2}Institute for Network Sciences and Cyberspace, Tsinghua University
}

\usepackage{bibentry}

\begin{document}

\maketitle

\begin{abstract}
Vision-and-Language Navigation (VLN) requires an agent to interpret natural language instructions and navigate complex environments. Current approaches often adopt a "black-box" paradigm, where a single Large Language Model (LLM) makes end-to-end decisions. However, it is plagued by critical vulnerabilities, including poor spatial reasoning, weak cross-modal grounding, and memory overload in long-horizon tasks. To systematically address these issues, we propose Memory Spatial Navigation(MSNav), a framework that fuses three modules into a synergistic architecture, which transforms fragile inference into a robust, integrated intelligence. MSNav integrates three modules: Memory Module, a dynamic map memory module that tackles memory overload through selective node pruning, enhancing long-range exploration; Spatial Module, a module for spatial reasoning and object relationship inference that improves endpoint recognition; and Decision Module, a module using LLM-based path planning to execute robust actions. Powering Spatial Module, we also introduce an Instruction-Object-Space (I-O-S) dataset and fine-tune the Qwen3-4B model into Qwen-Spatial (Qwen-Sp), which outperforms leading commercial LLMs in object list extraction, achieving higher F1 and NDCG scores on the I-O-S test set. Extensive experiments on the Room-to-Room (R2R) and REVERIE datasets demonstrate MSNav's state-of-the-art performance with significant improvements in Success Rate (SR) and Success weighted by Path Length (SPL).
\end{abstract}

\section{Introduction}

We hope that VLN agents can eventually achieve navigating complex environments with remarkable efficiency. For example, when instructed to ``reach the kitchen's refrigerator,'' an ideal VLN agent can visualize the kitchen layout, focus on key landmarks, and filter out irrelevant details from memory. However, existing LLM-based Zero-Shot VLN (ZS-VLN) solutions frequently falter in complex, long-distance tasks and scenarios with ambiguous endpoints \cite{zhou2023navgpt, long2023discuss, chen2024mapgpt}.

Inspired by the failures in prior VLN solutions, we propose Memory Spatial Navigation(MSNav), a framework together with dynamic memory, LLM spatial reasoning and LLM-based planning, which transforms fragile black-box inference into a robust, integrated intelligence. MSNav integrates three modules:(1) Memory Module, creating a dynamic topological map that selectively retains important spatial information while discarding outdated or irrelevant details; (2) Spatial Module, presenting reasonable spatial imagination and reasoning capabilities, analyzing instructions and infers spatial layouts from linguistic cues, and enhancing t he agent's ability to understand environmental context; and (3) Decision Module, advanced LLMs to determine navigation actions based on instructions, processed observations, object spatial layouts, and map information. 

Additionally, we construct an Instruction-Object-Space (I-O-S) dataset, derived from oracle paths across indoor environments, to support instruction analysis and spatial reasoning. We have also fine-tuned the Qwen3-4B model\cite{yang2025qwen3} on this I-O-S dataset to create the Qwen-Sp model, which can analyze language instructions for VLN tasks, extract and reason about objects along the navigation path, and infer spatial layouts of objects at the destination. Extensive experiments on the Room-to-Room (R2R) and REVERIE datasets \cite{anderson2018vision, qi2020reverie} demonstrate MSNav's state-of-the-art performance, with significant gains in Success Rate (SR) and Success weighted by Path Length (SPL).

Our contributions are:
\begin{itemize}
    \item We introduce MSNav, achieving state-of-the-art performance with a 5.1\% improvement in SR and 5.0\% in SPL on the R2R subset \cite{anderson2018vision}. 
    \item We present the I-O-S dataset, comprising 28,414 samples, enabling fine-grained analysis of navigation instructions. 
    \item We develop Qwen-Sp, outperforming leading commercial LLMs~\cite{openai2024gpt4o,google2025gemini25flash,xai2025grok3} in the task of object extraction, achieving a higher F1 score (0.316 vs. 0.270 for GPT-4o) and NDCG score (0.388 vs. 0.325 for GPT-4o) on the I-O-S test set. 
    \item We demonstrate the versatility of the Spatial Module, which can be seamlessly integrated into other VLN frameworks to enhance their performance, whether map-based or not.

\end{itemize}

\section{Related Work}

\noindent\textbf{Vision-and-Language Navigation}
Vision-and-Language Navigation (VLN) tasks an agent with following natural language instructions in 3D environments \cite{anderson2018vision, krantz2020beyond, chen2019touchdown, qi2020reverie}. Foundational research in VLN was dominated by supervised methods focusing on cross-modal alignment \cite{hao2020towards, hong-etal-2021-vln, chen2021hamt, chen2020uniter, li2020oscar} and was supplemented by techniques like data augmentation, specialized training, and self-correction \cite{fried2018speaker, tan-etal-2019-learning, wang2023scaling, wang2019reinforced, huang2019transferable, ke2019tactical, ma2019regretful}. A central component has always been spatial memory, implemented via metric maps \cite{thrun1998learning, fuentes2015visual} or efficient topological graphs to manage historical information and environmental structure \cite{chen2021history, deng2020evolving, chen2021topological, chen2022think}.

\noindent\textbf{Large Language Models in VLN}
Large Language Models (LLMs) \cite{brown2020language, openai2023gpt4} have introduced a new paradigm, enabling Zero-Shot VLN (ZS-VLN) where they serve as out-of-the-box planners. Seminal works like NavGPT \cite{zhou2023navgpt} demonstrated this feasibility, while subsequent approaches like DiscussNav \cite{long2023discuss} and MapGPT \cite{chen2024mapgpt} enhanced LLM reasoning by incorporating multi-agent dialogue or topological maps. In parallel, LLMs are also being successfully fine-tuned for navigation \cite{pan2023langnav, lin2024navcot}. Effective prompting remains crucial across all LLM-based methods to elicit strong performance \cite{wei2022chain, kojima2022large, yao2022react}.

\begin{figure*}[t]
  \centering
  \includegraphics[width=0.95\textwidth]{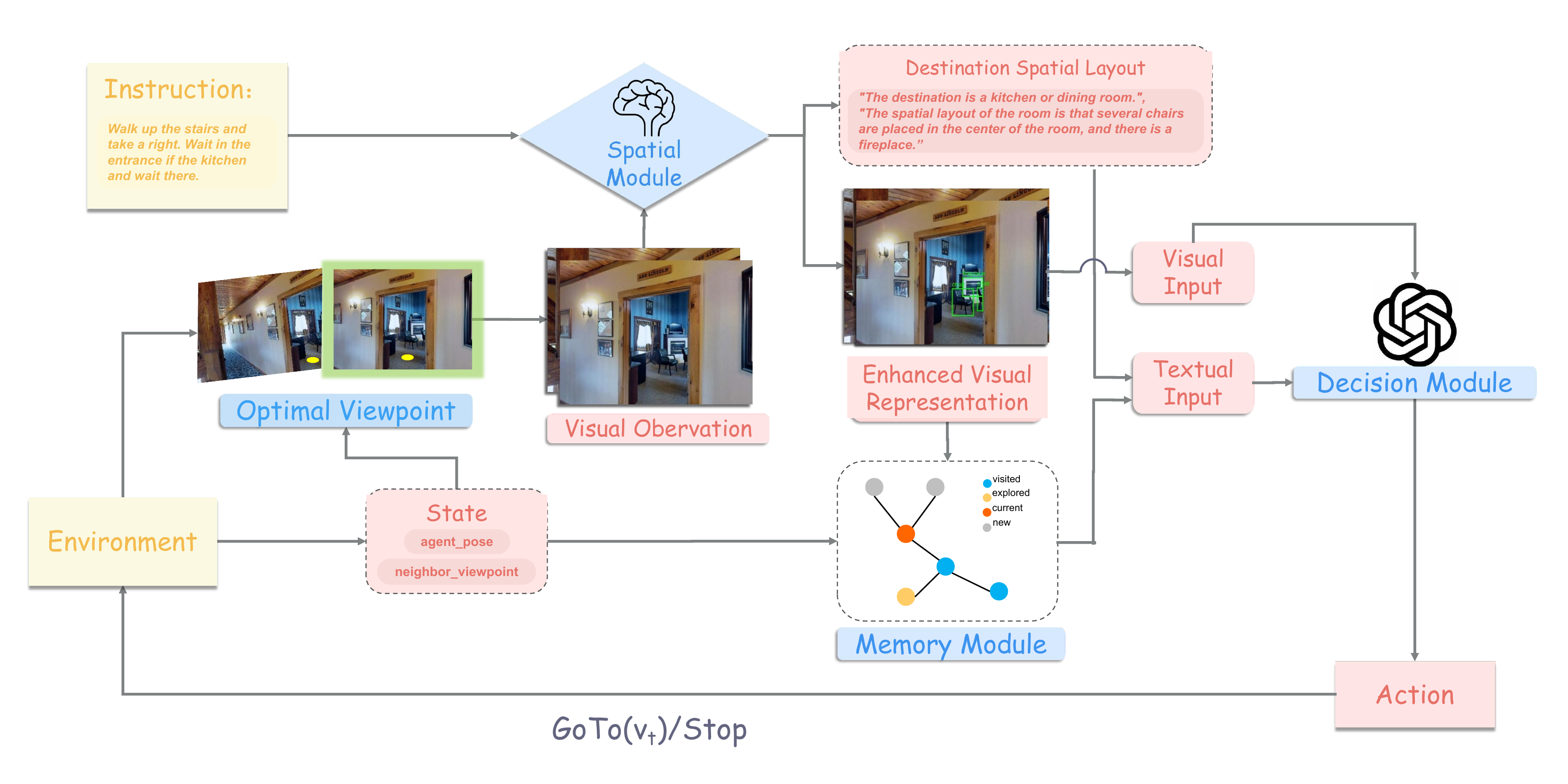}
  \caption{The MSNav architecture. The agent perceives the environment and then identifies optimal navigable viewpoint(yellow dot shows the best viewpoint that contains complete spatial information, explanation only). Memory Module maintains a topological map for long-term context. Spatial Module reasons about the target location by processing instructions and visual cues. Based on the map, layout, and current view, the Decision Module outputs navigation action(stop or proceed to a selected viewpoint), and the 'observe-reason-act' cycle repeats until task completion.}
  \label{fig:msnav}
\end{figure*}

\section{Methodology}
\noindent\textbf{Task Description} VLN tasks require an agent to interpret a natural language instruction \( I = \{w_1, w_2, \dots, w_L\} \) and navigate a 3D environment to a target location. At each step \( t \), given the current pose \( p_t \), the simulator provides several neighboring viewpoints that are currently navigable. The agent observes its state \( s_t \), including a set of navigable viewpoints \( \mathcal{V}_t = \{v_{t,i}\}_{i=1}^K \), where \( K \) is the number of navigable viewpoints, and visual observations \( O \), and selects an action \( a_t \) from a discrete action space \( A_t \) (e.g., navigate to an adjacent viewpoint or stop). The action is sent to the control module to execute the corresponding movement. The challenge lies in grounding linguistic instructions in visual scenes to generate an action sequence \( A = \{a_1, a_2, \dots, a_T\} \).

MSNav adopts a modular framework that integrates three key processes: memory, spatial reasoning, and decision-making. The framework comprises Memory Module, Spatial Module, and Decision Module, which work together to process visual and linguistic inputs efficiently, as illustrated in Figure~\ref{fig:msnav}. These modules have a mutual collaboration and cooperation to achieve robust and effective navigation in complex indoor environments.

\subsection{Memory Module}
\label{sec:msmem}

\begin{figure*}
  \centering
  \includegraphics[width=\textwidth, height=0.3\textheight, keepaspectratio]{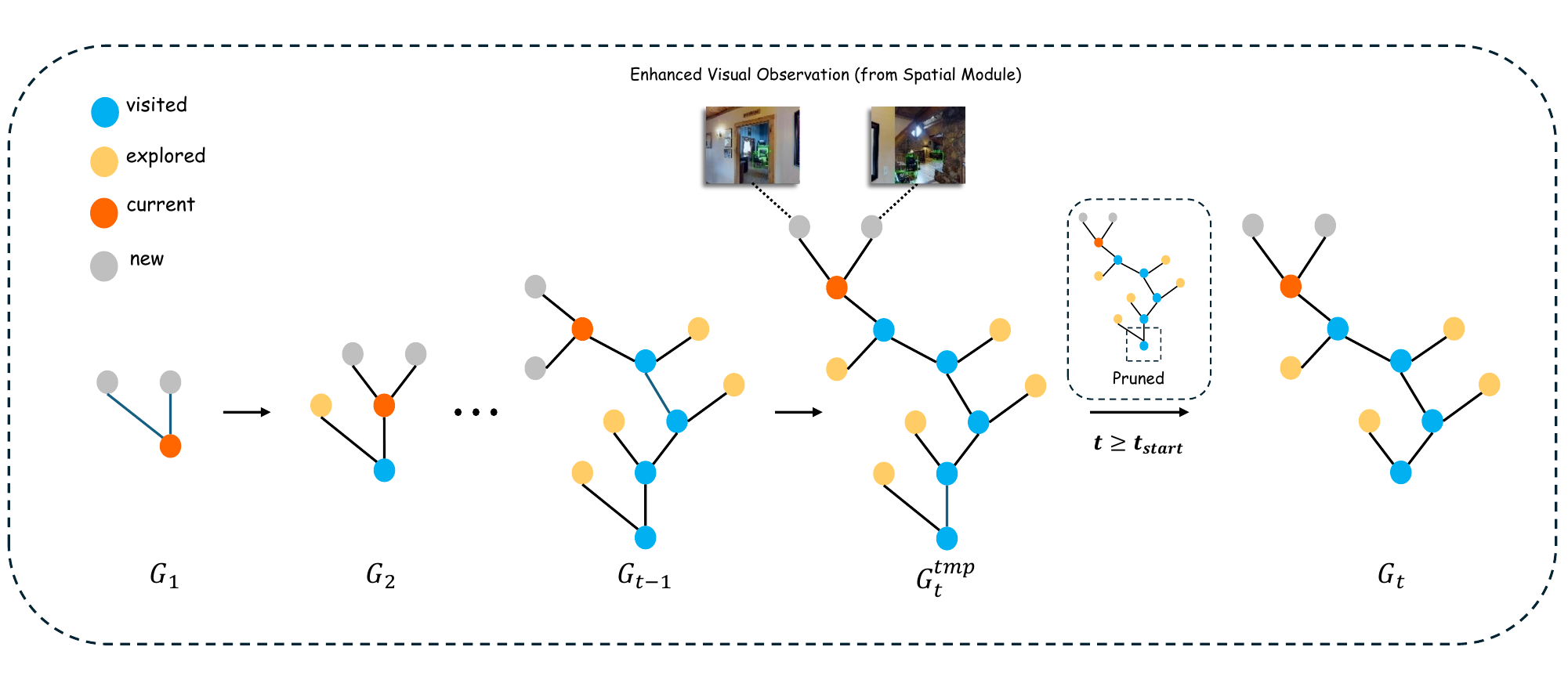}
  \caption{Memory Module, illustrating the dynamic construction and pruning of a task-relevant topological map. At step \( t \), the Memory Module observes navigable viewpoints and uses their enhanced visual observations as representations, adding them as new nodes (gray) to the previous map \( G_{t-1} \) to form an intermediate map \( G_t^{\text{tmp}} \); if \( t \geq t_{\text{start}} \), a pruning operation is triggered, removing \( N_{\text{remove}} \) nodes based on their pruning priority scores (\( N_{\text{remove}} = 1 \) in this figure), resulting in a compact map \( G_t \).}
  \label{fig:msmem}
\end{figure*}

Constructing navigation maps\citep{epstein2017cognitive} and a strong memory system\citep{baddeley1974, malleret2024} is proved helpful in completing a navigation task in a complex environment. We investigated current VLN performances, noting that existing methods usually fail in prolonged tasks\citep{zhou2023navgpt, long2023discuss, chen2024mapgpt}, particularly beyond 13 steps in the R2R dataset \citep{anderson2018vision}. Excessive node accumulation in topological maps overwhelms LLM context limits, reducing success rates. Unlike prior approaches that retain all observations in an expanding map (\citealp{chen2022think}; \citealp{chen2024mapgpt}), our Memory Module dynamically filters irrelevant nodes, sustaining performance and mitigating LLM context constraints. The overall workflow of Memory Module is illustrated in Figure~\ref{fig:msmem}.

\subsubsection{Map Construction}
In VLN, the agent builds a real-time map of an unfamiliar environment using observations from exploration. Following prior work (\citealp{chen2022think}; \citealp{chen2024mapgpt}), we use a topological graph \( G_t = (V_t, E_t) \), where \( V_t = \{v_{t,i}\}_{i=1}^K \) represents viewpoint nodes observed up to time step \( t \), and \( E_t \) denotes navigable connections between them.

At each step \( t \), the agent records new viewpoints and their connections based on the simulator’s feedback about neighboring nodes. These are added to an intermediate graph \( G_t^{\text{tmp}} \), updated from the previous graph \( G_{t-1} \). After obtaining the intermediate graph \( G_t^{\text{tmp}} \), it is dynamically pruned to produce the final graph \( G_t \).

\subsubsection{Dynamic Map Pruning}
To maintain a compact and task-relevant topological map, Memory Module dynamically evaluates and prunes nodes from the intermediate graph \( G_t^{\text{tmp}} \) that are no longer pertinent to the navigation task. This process begins after an initial exploration phase (\( t \geq t_{\text{start}} \)), ensuring the map remains efficient by removing outdated or irrelevant information. By selectively filtering nodes, the Memory Module reduces memory overhead and mitigates interference from obsolete data, producing the final graph \( G_t \).

The set \( T_t \subseteq V_t^{\text{tmp}} \) represents all viewpoint nodes that the agent has visited up to time step \( t \). This set tracks the agent's exploration history and is used to assess node relevance.

The Memory Module identifies a subset of nodes \( \mathcal{V}_{\text{assess}} \subseteq V_t^{\text{tmp}} \) for relevance evaluation based on three criteria: nodes must be \textit{non-current}, meaning they are not the agent’s current viewpoint (\( v_{t,i} \neq v_t \)); they must be \textit{previously visited}, having been explored (\( v_{t,i} \in T_t \)); and they must be \textit{temporally stale}, not revisited recently, satisfying \( t - \tau(v_{t,i}) > \theta_{\text{recent-visit}} \) and \( t - \tau(v_{t,i}) > \theta_{\text{age}} \), where \( \tau(v_{t,i}) \) is the time step when node \( v_{t,i} \) was last visited, \( \theta_{\text{recent-visit}} \) is the minimum time elapsed since the last visit to consider a node for pruning, and \( \theta_{\text{age}} \) is the threshold for determining node staleness based on its age.

Nodes in \( \mathcal{V}_{\text{assess}} \) are assigned a pruning priority score \( P(v_{t,i}) \), which quantifies their relevance to the ongoing task:

\begin{align}
P(v_{t,i}) = {} & \lambda_t f_t(v_{t,i}) + \lambda_d f_d(v_{t,i}) \notag \\
               & + \lambda_f f_f(v_{t,i}) + \lambda_{\text{dist}} f_{\text{dist}}(v_{t,i})
\label{eq:priority_score}
\end{align}

where:
\begin{itemize}
    \item \( f_t(v_{t,i}) = \max(1, t - \tau(v_{t,i}) - \theta_{\text{age}}) \): Measures temporal staleness, prioritizing older nodes.
    \item \( f_d(v_{t,i}) = -\deg_{G_t^{\text{tmp}}}(v_{t,i}) \): prioritizes nodes with high connectivity, as they are more critical to navigation.
    \item \( f_f(v_{t,i}) = -|\{v_{t,j} \mid (v_{t,i}, v_{t,j}) \in E_t^{\text{tmp}} \land v_{t,j} \notin T_t\}| \): prioritizes nodes with fewer unexplored neighbors, indicating lower exploration potential.
    \item \( f_{\text{dist}}(v_{t,i}) = d_{G_t^{\text{tmp}}}(v_t, v_{t,i}) \): Considers the graph distance from the current viewpoint, prioritizing distant nodes.
\end{itemize}
The coefficients \( \lambda_t, \lambda_d, \lambda_f, \lambda_{\text{dist}} \) balance the contributions of each factor. A higher pruning priority value indicates a greater possibility of pruning.

Based on the pruning priority scores, the top \( N_{\text{remove}} \) nodes with the highest \( P(v_{t,i}) \) are removed from \( G_t^{\text{tmp}} \), yielding the final map \( G_t = (V_t, E_t) \). This selective pruning ensures that the topological map remains concise, relevant, and computationally efficient, supporting robust navigation over extended periods.

\subsubsection{Map Representation}
\label{sec:Memory Module_llm_presentation}
The Memory Module structures the filtered topological map \( G_t = (V_t, E_t) \) into prompts for Decision Module, adapting insights from prior work~\cite{chen2024mapgpt}. The prompts include: (1) Trajectory, listing visited node identifiers in \( V_t \); (2) Map, detailing node connectivity in \( E_t \); and (3) Supplementary Information, linking nodes \( v_t \) to enhanced visual observations \( O_{k^*}^e \) from Spatial Module. This ensures a concise, task-relevant spatial representation for the LLM. Detailed prompts are in Appendix.

\subsection{Spatial Module}
\label{sec:msspace}
\begin{figure}[!htbp]
  \centering
  \includegraphics[width=0.95\linewidth]{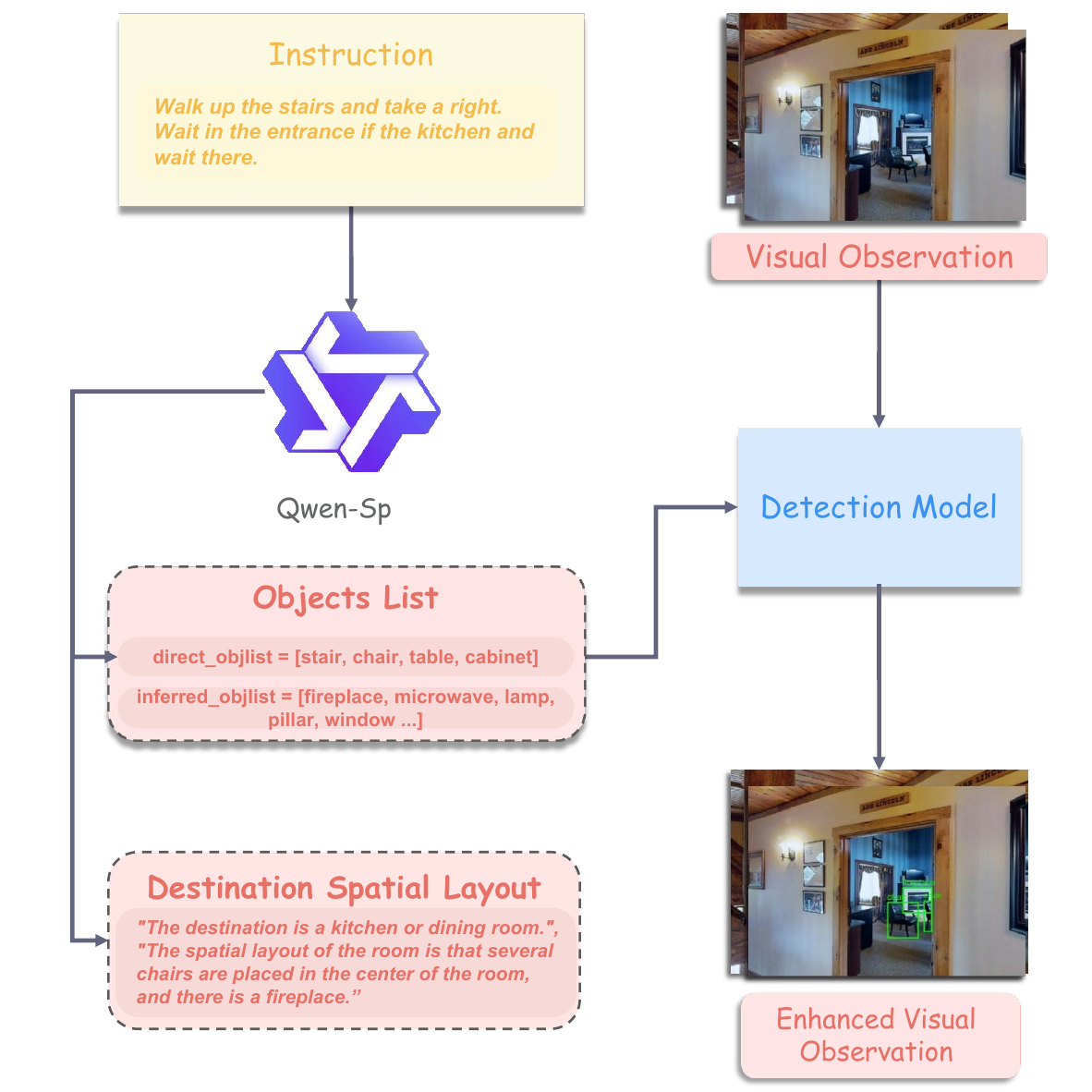}
  \caption{Spatial Module architecture depicts the pipeline for spatial reasoning. Qwen-Sp processes instructions to extract object lists and infer destination spatial layouts, while YOLO-World as detection model, enhances visual observations to highlight task-relevant objects for the agent. }
  \label{fig:msspace}
\end{figure}

In navigation tasks, landmarks and spatial configurations at destinations contain key information to achieve precise location identification. In VLN, agents often misidentify targets, such as stopping in a hallway instead of a kitchen in R2R tasks \citep{anderson2018vision} due to weak spatial reasoning. Spatial Module addresses this by extracting task-relevant objects from instructions and inferring destination layouts. Using our I-O-S dataset, we fine-tuned Qwen3-4B into Qwen-Sp to generate accurate object lists and layouts, boosting navigation precision. The Spatial Module is shown in Figure~\ref{fig:msspace}. Beyond VLN, the I-O-S dataset also enhances LLMs’ spatial imagination and reasoning capabilities.

\subsubsection{I-O-S Dataset}
The Instruction-Object-Space (I-O-S) dataset is a novel resource designed to enhance spatial reasoning in VLN by providing structured data that links natural language instructions to objects and their spatial arrangements. Comprising 28,414 samples derived from expert trajectories in indoor environments(25,694 training samples and 2,720 test samples), the I-O-S dataset captures three key components: (1) Instructions, which are natural language navigation directives; (2) Objects, a list of task-relevant objects encountered along the trajectory or at the destination; and (3) Destination Spatial Layouts, describing the relative positions of objects at the destination (e.g., ``The spatial layout of the room is that several chairs are
placed in the center of the room, and there is a fireplace''). Each sample in the I-O-S dataset is formatted as a tuple \( (I, O, S) \), where \( I \) is the instruction, \( O \) is the set of objects, and \( S \) is the description of destination spatial layout.

To construct the dataset, we extracted oracle paths from indoor environments, which provide optimal navigation trajectories. Objects along the path and at the destination were identified using the simulator’s ground-truth object annotations(including object lists and their bounding box coordinates). Spatial arrangements were generated through a two-step process: first, an LLM proposed candidate layouts based on object lists and their bounding box coordinates observed in the destination scenes; second, human annotators verified and refined these layouts to ensure accuracy and consistency through iterative subset sampling. By providing fine-grained annotations, the I-O-S dataset enables models to learn extracting task-relevant objects from instructions and inferring their spatial configurations. See Appendix for details.

\subsubsection{Spatial Reasoning Model}
To enable robust spatial reasoning, we developed Qwen-Sp by fine-tuning Qwen3-4B \cite{yang2025qwen3} on the I-O-S dataset using Low-Rank Adaptation (LoRA) \cite{hu2022lora}. Qwen-Sp employs two LoRA adapters: one to extract task-relevant objects from navigation instructions (e.g., identifying ``refrigerator'' from ``go to the kitchen’s refrigerator'') and another to infer their spatial arrangements at the destination (e.g., ``the refrigerator is against the kitchen’s back wall''). Fine-tuned on 25,694 I-O-S samples, Qwen-Sp achieves superior performance in object list extraction compared to leading commercial LLMs, including GPT-4o, Gemini-2.5-Flash, and Grok3 \cite{xai2025grok3}, as detailed in Section Experiments. This highlights Qwen-Sp’s ability to accurately identify and prioritize task-relevant objects, which is critical for effective navigation. For zero-shot REVERIE experiments \cite{qi2020reverie}, we avoid direct use of the fine-tuned model, instead leveraging its learned patterns to design prompts for commercial LLMs (e.g., GPT-4o). Qwen-Sp with training details is provided in the Appendix.

\subsubsection{Visual Input Enhancement}

The Visual Input Enhancement component enables landmark-based pathfinding by enhancing visual observations with task-relevant objects from the spatial reasoning model’s object list, highlighting landmarks for navigation \cite{skaramagkas2021eye}.

We first compute the target direction $(\theta_{\text{tg}}, \phi_{\text{tg}})$ from the agent's current pose $p_t$, and selects the information-rich view visual observation $O_{k^*}$ by finding the camera view direction $(\theta_k, \phi_k)$ that minimizes the $L_1$ angular distance to this target:
\begin{equation}
k^* = \underset{k \in \text{available views}}{\arg\min}\mathcal{D}((\theta_k, \phi_k), (\theta_{\text{tg}}, \phi_{\text{tg}})).
\end{equation}

Then, the component uses YOLO-World \cite{cheng2024yolo}, a lightweight, fast, and high-performance open-vocabulary object detection system, to annotate objects in the visual observation \( O_{k^*} \) selected, guiding the agent along the instruction-specified path. The enhanced viewpoint observation ensures that the representation encapsulates sufficient spatial information and visual features for effective navigation.

\subsection{Decision Module}
\label{sec:msdecision}

\begin{figure*}[t]
  \centering
  \includegraphics[width=0.95\textwidth]{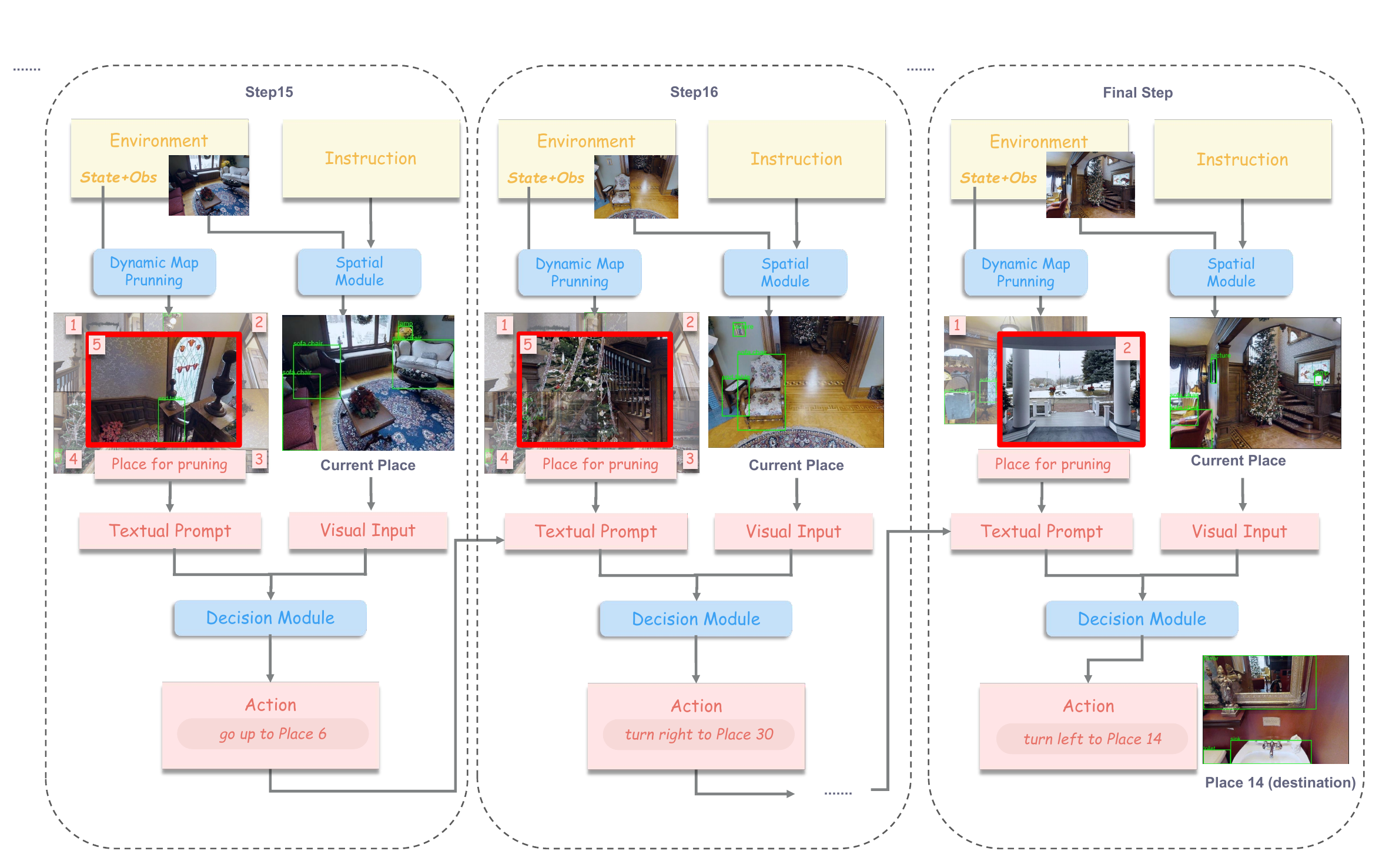}
  \caption{MSNav's iterative VLN process (key steps shown). The Memory Module dynamically prunes environment states, forming a map for context. The Spatial Module (via Qwen-Sp) processes instructions and visual observations, inferring object lists for visual enhancement and destination spatial layouts. The Decision Module then uses this integrated textual prompt and enhanced visual input to make action decisions, repeating the cycle until task completion.}
  \label{fig:case_study}
\end{figure*}

To harness reasoning and decision-making capabilities, we introduce the Decision Module, a module that employs an advanced LLM, GPT-4o, to facilitate high-level decision-making. At each time step \( t \), Decision Module processes the following inputs: the natural language instruction (\( I \)), specifying the navigation goal; the Memory Module context (\( C_t^{\text{Memory}} \)), including the trajectory and the map; the destination spatial layout (\(\ SL^{\text{Spatial}} \)) provided by Spatial Module; and other prompt information (e.g., history, previous planning, and action options, denoted as \( P_t \)). The LLM integrates these inputs to output an action \( a_t \), which is either the selection of a neighboring viewpoint or a decision to stop:
\begin{equation}
a_t = \text{Decision}(I, C_t^{\text{Memory}}, SL^{\text{Spatial}}, P_t).
\end{equation}
Our prompt design draws on insights from prior work~\cite{chen2024mapgpt} while incorporating adaptations tailored to MSNav's framework. Full prompt structure is detailed in the Appendix. The task workflow for MSNav is shown in the Figure~\ref{fig:case_study}.

\section{Experiments}
\label{sec:experiments}

\subsection{Experimental Settings}
MSNav is evaluated on the R2R \cite{anderson2018vision} and REVERIE \cite{qi2020reverie} datasets, which are widely used benchmarks in zero-shot VLN setting, with the I-O-S dataset used to assess Qwen-Sp’s spatial inferring ability. MSNav is compared to NavGPT \cite{zhou2023navgpt}, DiscussNav \cite{long2023discuss}, and MapGPT \cite{chen2024mapgpt}, using GPT-4o for a fair comparison. Qwen-Sp is also tested against GPT-4o, Gemini-2.5-Flash, and Grok3 on the I-O-S dataset. We conduct VLN experiments on the Matterport3D simulator\cite{chang2017matterport3d}. The implementation details are in the Appendix.

\noindent\textbf{Evaluation Metrics}
Performance is assessed using the following metrics. For VLN tasks: (1) \textit{Success Rate (SR)}, the percentage of successful episodes; (2) \textit{Success weighted by Path Length (SPL)}, which balances success and path efficiency; (3) \textit{Oracle Success Rate (OSR)}, the SR with an oracle stop policy; and (4) \textit{Navigation Error (NE)}, the average distance in meters to the target. For evaluating the spatial inference capabilities of LLMs: (5) \textit{F1 Score}, measuring precision and recall for object list extraction; and (6) \textit{Normalized Discounted Cumulative Gain (NDCG)}, assessing the ranking quality of extracted objects. Details for the standard metrics (1-6) are in the Appendix. 

Additionally, we introduce a novel metric, \textit{Map Efficiency (ME)}, which evaluates the quality of topological maps in VLN tasks. It is defined as:
\begin{equation}
\text{ME} = \frac{|T_t \cap T_{\text{expert}}|}{|T_{\text{expert}}|} \cdot \frac{1}{1 + \alpha \cdot \frac{|V_t|}{|T_{\text{expert}}|}},
\end{equation}
where ME balances the coverage of expert path nodes (\(T_{\text{expert}}\)) by the agent's trajectory (\(T_t\)) against the total map size (\(V_t\)), rewarding compact yet accurate maps. Step by step, we deeply analyze the performance of Memory Module's node discard selection during task execution, and ultimately selected the penalty factor \(\alpha=0.25\).

\subsection{Experimental Results}
\noindent\textbf{ZS-VLN Benchmark Comparison}
Following prior work \cite{zhou2023navgpt, chen2024mapgpt}, we evaluate MSNav on the standard R2R subset consisting of 72 scenes and 216 samples(we name it as R2R-ZS uniformly), as shown in Table~\ref{tab:r2r_subset}. MSNav achieves an SR of 50.9\% and SPL of an 42.6\%, outperforming MapGPT by 5.1\% and 5.0\%, respectively. Memory Module’s pruning maintains compact maps, with an ME of 40.4\% , enabling stable exploration in long trajectories. Notably, MSNav’s higher OSR (7.3\% higher than MapGPT) likely stems from Memory Module’s pruning, facilitating late-stage exploration without increased resource demands.

\begin{table}[!htbp]
\centering
\small
\begin{tabular}{l@{\hspace{0pt}}c@{\hspace{0pt}}c@{\hspace{0pt}}c@{\hspace{0pt}}c@{\hspace{0pt}}c@{\hspace{0pt}}}
\toprule
Methods & SR$\uparrow$ & SPL$\uparrow$ & OSR$\uparrow$ & NE$\downarrow$ & ME$\uparrow$ \\
\midrule
NavGPT \cite{zhou2023navgpt} & 36.1 & 31.6 & 40.3 & 6.26 & - \\
DiscussNav \cite{long2023discuss} & 37.5 & 33.3 & 51.0 & 6.30 & - \\
MapGPT \cite{chen2024mapgpt} & 45.8 & 37.6 & 56.5 & 5.31 & 38.0 \\
MSNav (Ours) & \textbf{50.9} & \textbf{42.6} & \textbf{63.9} & \textbf{5.02} & \textbf{40.4} \\ 
\bottomrule
\end{tabular}
\caption{Comparison of ZS-VLN performance on the R2R-ZS. NavGPT and MapGPT results are reproduced using GPT-4o to ensure a fair comparison.}
\label{tab:r2r_subset}
\end{table}

\noindent\textbf{R2R Large-Scale Evaluation}
To compare with prior VLN and ZS-VLN work, we evaluate MSNav on the R2R full validation unseen set (11 scenes, 783 samples), as shown in Table~\ref{tab:fullr2r}. MSNav achieves an SR of 46\% and an SPL of 40\%, surpassing MapGPT by 2\% and 5\%, respectively. The relatively moderate improvements observed here can be attributed to the limited scene diversity within the 11-scene subset, which restricts the effectiveness of Memory Module’s pruning and Spatial Module’s spatial reasoning capabilities. However, MSNav’s SR outperforms three trained and pretrained methods, achieving state-of-the-art zero-shot performance.

\begin{table}[!htbp]
\centering
\small
\begin{tabular}{l@{\hspace{0pt}}l@{\hspace{1pt}}c@{\hspace{0pt}}c@{\hspace{0pt}}c@{\hspace{0pt}}c}
\toprule
Settings & Methods & SR & SPL & OSR & NE \\
\midrule
\multirow{3}{*}{Train}
& Seq2Seq~\cite{anderson2018vision} & 21 & - & 28 & 7.81 \\
& Speaker~\cite{fried2018speaker} & 35 & - & 45 & 6.62 \\
& EnvDrop~\cite{tan-etal-2019-learning}& 52 & 48 & - & 5.22 \\
\midrule
\multirow{8}{*}{Pretrain}
& PREVALENT \cite{hao2020towards} & 58 & 53 & - & 4.71 \\
& RecBERT \cite{hong2021vlnbert}& 63 & 57 & 69 & 3.93 \\
& HAMT \cite{chen2021history} & 66 & 61 & 73 & 2.29 \\
& DUET \cite{chen2022think} & 72 & 60 & 81 & 3.31 \\
& LangNav \cite{pan2023langnav} & 43 & - & - & - \\
& ScaleVLN \cite{wang2023scaling} & 81 & 70 & 88 & 2.09 \\
& NavCoT \cite{lin2024navcot} & 40 & 37 & 48 & 6.26 \\
& NaviLLM \cite{zheng2024learninggeneralistmodelembodied}& 67 & 59 & - & 3.51 \\
& \makecell[l]{NavGPT-2$_{\text{FlanT5-XL}}$ \\ \cite{zhou2024navgpt2unleashingnavigationalreasoning}} & 68 & 56 & 74 & 3.37 \\
\midrule
\multirow{4}{*}{ZS}
& NavGPT \cite{zhou2023navgpt} & 34 & 29 & 42 & 6.46 \\
& DiscussNav \cite{long2023discuss} & 43 & 40 & 61 & 5.32 \\
& MapGPT \cite{chen2024mapgpt}  & 44 & 35 & 58 & 5.63 \\
& \textbf{MSNav (Ours)}  & \textbf{46} & \textbf{40} & \textbf{65} & \textbf{5.24} \\
\bottomrule
\end{tabular}
\caption{Performance comparison on the complete validation unseen set of the R2R dataset (11 scenes, 783 samples). MSNav achieves the highest SR among all zero-shot methods and surpasses three trained and pretrained approaches. }
\label{tab:fullr2r}
\end{table}

\noindent\textbf{REVERIE Complex Task Evaluation}
In line with the methodology of prior benchmarks \cite{chen2024mapgpt}, we evaluate MSNav on the randomly sampled REVERIE subset (70 scenes, 140 samples, named REVERIE-ZS), as shown in Table~\ref{tab:reverie}. we used GPT-4o in the Spatial Module for a true zero-shot evaluation, avoiding I-O-S data contamination. MSNav achieves an SR of 45.7\% and an SPL of 32.8\%, surpassing MapGPT by 4.3\% and 4.4\%, respectively. 
This success highlights the Memory Module's strength in handling REVERIE's complex, long-range tasks, while the Spatial Module's object extraction and the viewpoint selection are also contributed. The limited ME improvement is likely due to REVERIE's rich instructions aiding map construction across all methods.

\begin{table}[!htbp]
\centering
\small
\begin{tabular}{l@{\hspace{1pt}}l@{\hspace{0pt}}c@{\hspace{1pt}}c@{\hspace{1pt}}c@{\hspace{1pt}}c@{\hspace{1pt}}c@{\hspace{0pt}}}
\toprule
Settings & Methods & SR & SPL  & OSR  & NE   & ME \\
\midrule
\multirow{6}{*}{Train}
& Seq2Seq~\cite{anderson2018vision} & 4.2 & 2.84 & 8.07 & - & - \\
& Airbert~\cite{guhur2021airbertindomainpretrainingvisionandlanguage} & 27.9 & 21.9 & 34.5 & - & - \\
& TD-STP~\cite{zhao2022targetdrivenstructuredtransformerplanner} & 34.9 & 27.3 & 39.5 & - & - \\
& SIA~\cite{lin2021sceneintuitiveagentremoteembodied} & 31.5 & 16.3 & 44.7 & - & - \\
& DUET~\cite{chen2022thinkglobalactlocal} & 46.9 & 33.7 & 51.0 & - & - \\
& AutoVLN~\cite{chen2022learningunlabeled3denvironments}& 55.9 & 40.8 & 62.1 & - & - \\
\midrule
\multirow{3}{*}{ZS}
& \makecell[l]{NavGPT \\ \cite{zhou2023navgpt}} & 28.9 & 23.0 & 32.6 & 7.86 & - \\
& MapGPT \cite{chen2024mapgpt} & 41.4 & 28.4 & 56.4 & 7.12 & 34.7 \\
& MSNav (Ours) & \textbf{45.7} & \textbf{32.8} & \textbf{59.3} & \textbf{7.89} & \textbf{35.6} \\
\bottomrule
\end{tabular}
\caption{Comparison of performance on the REVERIE val unseen dataset. Models in the ``Train'' setting are evaluated on the full set, while Zero-shot (ZS) models are evaluated on REVERIE-ZS. For a fair comparison, results for NavGPT and MapGPT are our reproductions using GPT-4o. }
\label{tab:reverie}
\end{table}

\noindent\textbf{LLM Spatial Inference Comparison}
Our evaluation on the I-O-S test set(2720 samples), as shown in Table~\ref{tab:qwen_sp_comparison}, demonstrates that language models can be effectively trained for spatial reasoning. The fine-tuned Qwen-Sp, with a leading F1 score of 0.316 and an NDCG of 0.388, significantly outperforms a strong one-shot baseline like GPT-4o (+0.046 and +0.063, respectively). This success underscores that targeted fine-tuning is a viable path for imbuing language models with specialized spatial intelligence, presenting a promising direction for future research. Although the indoor-centric I-O-S dataset currently limits generalization, extending this training approach to more diverse environments, such as outdoor scenes, is a key area for future work.

\begin{table}[!htbp]
\centering
\small
\begin{tabular}{l|cccc}
\toprule
Model & F1DO $\uparrow$ & F1IO $\uparrow$ & F1 $\uparrow$ & NDCG $\uparrow$ \\
\midrule
GPT-4o & 0.258 & 0.150 & 0.270 & 0.325 \\
Grok3 & 0.055 & 0.057 & 0.096 & 0.095 \\
Gemini-2.5-Flash & 0.023 & 0.055 & 0.096 & 0.106 \\
Qwen3-4B & 0.236 & 0.039 & 0.138 & 0.198 \\
Qwen-Sp (Ours) & \textbf{0.357} & \textbf{0.179} & \textbf{0.316} & \textbf{0.388} \\
\bottomrule
\end{tabular}
\caption{Comparative evaluation of object extraction capabilities of different LLMs on the I-O-S test set (2,720 samples). F1DO and F1IO represent the F1 scores for direct and inferred objects, respectively. Qwen-Sp outperforms other models across all metrics.}
\label{tab:qwen_sp_comparison}
\end{table}

\subsection{Ablation Study}
As shown in Table~\ref{tab:ablation}, our ablation studies on R2R-ZS confirm the effectiveness of each MSNav module. Notably, removing the Memory Module significantly reduces OSR, which is consistent with its design for improving long-range exploration under oracle stopping conditions. Furthermore, the Spatial Module demonstrates seamless transferability. When integrated into diverse map-based and non-map-based frameworks (e.g., applying its Destination Spatial Layout to NavGPT) via simple prompt modifications, it yields substantial improvements, confirming its effectiveness as a pluggable component.

\subsection{Long-Distance Analysis and Efficiency}

\subsubsection{Performance in Long Distance Tasks}
On 25 identified long-distance tasks (exceed 10 steps) from the R2R-ZS experiment, MSNav achieved 11 successes versus MapGPT's 6, increasing the success rate by 20.0\% (details in Table~\ref{tab:long_horizon_comparison}). The higher success rate of MSNav in long-distance tasks is mainly attributed to the dynamic pruning of the Memory Module, which discards outdated and irrelevant information, enabling the LLM to focus on the latest data and make accurate judgments under extended trajectories.

\begin{table}[htbp]
\centering
\small
\begin{tabular}{l@{\hspace{0pt}}|c@{\hspace{0pt}}c@{\hspace{0pt}}c@{\hspace{0pt}}c}
\toprule
Methods & SR$\uparrow$ & SPL$\uparrow$ & OSR$\uparrow$ & NE$\downarrow$ \\
\midrule
NavGPT \cite{zhou2023navgpt} & 36.1 & 31.6 & 40.3 & 6.26 \\
NavGPT+Spatial & 38.9 & 34.1 & 43.1 & 5.96 \\
MapGPT \cite{chen2024mapgpt} & 45.8 & 37.6 & 56.5 & 5.31 \\
MapGPT+Spatial & 48.1 & 39.6 & 58.3 & 5.11 \\
MSNav w/o Viewpoint Selection & 49.5 & 41.4 & 62.5 & 5.17 \\
MSNav w/o Memory & 48.1 & 40.1 & 58.8 & 5.32 \\
MSNav w/o Spatial & 47.7 & 39.6 & 61.1 & 5.37 \\
MSNav  & \textbf{50.9} & \textbf{42.6} & \textbf{63.9} & \textbf{5.02} \\
\bottomrule
\end{tabular}
\caption{Ablation study on R2R-ZS. The table shows the impact of removing MSNav's modules and the gains from integrating our Spatial Module into other frameworks. For NavGPT, only the Destination Spatial Layout was used.}
\label{tab:ablation}
\end{table}

\begin{table}[!htbp]
\centering
\small
\begin{tabular}{l@{\hspace{0pt}}c@{\hspace{0pt}}c@{\hspace{0pt}}c@{\hspace{0pt}}c@{\hspace{0pt}}c@{\hspace{0pt}}c@{\hspace{0pt}}c@{\hspace{0pt}}}
\toprule
\textbf{Method} & \textbf{Steps}$\downarrow$ & \textbf{Length}$\downarrow$ & \textbf{SR}$\uparrow$ & \textbf{SPL}$\uparrow$ & \textbf{OSR}$\uparrow$ & \textbf{NE}$\downarrow$ \\
\midrule
\makecell[l]{MapGPT \\ \cite{chen2024mapgpt}} & 11.84 & 22.12 & 24.0 & 10.0 & 48.0 & 10.69 \\
\textbf{MSNav(Ours)} & \textbf{9.24} & \textbf{15.40} & \textbf{44.0} & \textbf{29.0} & \textbf{60.0} & \textbf{5.41} \\
\bottomrule
\end{tabular}
\caption{Performance comparison between MapGPT and MSNav on 25 selected challenging long-horizon cases. MSNav demonstrates significant improvements across all key metrics.}
\label{tab:long_horizon_comparison}
\end{table}

\subsubsection{Efficiency and Cost Analysis}
Beyond navigation accuracy, we analyzed MSNav's computational efficiency. Processing a sample costs \$0.10–\$0.14 and takes 3–5 minutes in our experiments. This is driven by prompt lengths starting at 900–1,000 tokens and growing by 100–250 tokens per step. We consider this a reasonable trade-off for better zero-shot performance. Furthermore, our Memory Module mitigates this cost escalation by dynamically pruning the map, ensuring a manageable prompt size unlike prior methods that accumulate all information.

\section{Conclusion}
This paper introduces MSNav, a novel zero-shot vision-and-language navigation (ZS-VLN) framework that enhances navigation with dynamic memory, LLM spatial reasoning and LLM-based planning. MSNav demonstrates state-of-the-art performance on the R2R and REVERIE datasets, and its Spatial Module offers versatile plug-and-play integration to augment existing VLN frameworks. To further advance spatial understanding in large models, we introduced the Instruction-Object-Space (I-O-S) dataset. Leveraging this resource, we fine-tuned Qwen3-4B to develop Qwen-Sp, a model that demonstrably surpasses leading commercial LLMs like GPT-4o in critical instruction analysis and object extraction tasks.

\bibliography{custom}


\end{document}